\documentclass[journal]{IEEEtran}
\usepackage{amsmath,amsfonts}
\usepackage{diagbox}
\usepackage{booktabs}
\usepackage{algorithmic}
\usepackage{algorithm}
\usepackage{float}
\usepackage{array}
\usepackage[caption=false,font=normalsize,labelfont=sf,textfont=sf]{subfig}
\usepackage{textcomp}
\usepackage{stfloats}
\usepackage{url}
\usepackage{verbatim}
\usepackage{graphicx}
\usepackage{cite}
\begin{document}

\title{Partial Multi-View Clustering via Meta-Learning and Contrastive Feature Alignment}

\author{BoHao Chen
\thanks{}
\thanks{}}


\maketitle

\begin{abstract}
Partial multi-view clustering (PVC) presents significant challenges practical research problem for data analysis in real-world applications, especially when some views of the data are partially missing. Existing clustering methods struggle to handle incomplete views effectively, leading to suboptimal clustering performance. In this paper, we propose a novel dual optimization framework based on contrastive learning, which aims to maximize the consistency of latent features in incomplete multi-view data and improve clustering performance through deep learning models. By combining a fine-tuned Vision Transformer and k-nearest neighbors (KNN), we fill in missing views and dynamically adjust view weights using self-supervised learning and meta-learning. Experimental results demonstrate that our framework outperforms state-of-the-art clustering models on the BDGP and HW datasets, particularly in handling complex and incomplete multi-view data.
 
\end{abstract}

\section{Introduction}
\IEEEPARstart{M}{ultiview} data encompasses diverse features from different perspectives, such as sensors, multimodal inputs, observation points, and sources. Nowadays, multiview data has become quite common in practice due to the availability of rich multimedia data sampling equipment. For example, a video can contain image views, text views, and voice views (Wang et al., 2023a). However, the challenge of processing multiview data arises from the lack of available labels. A fundamental solution is to apply multiview clustering (MVC) (Li et al., 2019; Wen et al., 2022; Xu et al., 2022) to separate unlabeled multiview data into different clusters. Data within the same cluster is highly likely to belong to the same group while also belonging to the same category. Due to its expected performance, multiview clustering has been extensively studied, and numerous methods for multiview clustering (Tao et al., 2020; Zhang et al., 2017; Jiang et al., 2022) have been proposed. 

\section{Related Work}
In this section, we briefly review the related cross-modal clustering and incomplete cross-modal clustering.
\subsection{Multiview Clustering}
Traditional multiview clustering approaches are based on subspace learning, graph learning, spectral learning, and non-negative matrix factorization. For instance, Gao et al. (2015) proposed multiview subspace clustering (MVSC), which exhibits clustering based on consistent structures learned from the subspaces of each view. Wang et al. (2018b) developed multiview spectral clustering that utilizes low-rank matrix decomposition. Despite their effectiveness in handling multiview data, these methods capture the complex characteristics of multiview data. 

To overcome their limitations and inspired by deep learning, deep multiview clustering approaches (Chen et al., 2023; Wang et al., 2020a) have explored capturing nonlinear relationships in multiview data. For example, Li et al. (2019) designed deep multiview clustering via adversarial networks, while Andrew et al. (2013) developed a deep canonical correlation analysis (CCA) method for cross-view data.
\subsection{Partial Multiview Clustering}
Despite the successful outcomes of multiview approaches, they struggle to perform well on multiview data with missing one or two view information, particularly in incomplete/partial multiview data (Xu, Tao, and Xu, 2015; Wang et al., 2021; Zhang et al., 2020). This limitation complicates the application of these methods in real-world scenarios where data loss occurs due to sensor failures, noise, or transmission losses. To address this issue, several partial multiview clustering (PVMC) and incomplete multiview clustering (IMC) methods (Liu et al., 2018; Shao et al., 2016) have been proposed. The original partial multiview clustering methods were designed specifically for two views (Zhao, Liu, and Fu, 2016; Wang et al., 2018a, 2023b). For example, Li, Jiang, and Zhou (2014) used non-negative matrix factorization (NMF) and l-norm regularization to obtain complete subspaces for partial dual-view clustering, named PVC. To effectively handle incomplete multiview data with two or more views, Shao, He, and Philip (2015) developed various incomplete view clustering methods via weighted non-negative matrix factorization with l2,1-norm regularization, while Wen, Xu, and Liu (2018) proposed adaptive graph learning (IMSC). These methods have made significant advancements in partial multiview clustering.

Furthermore, to deeply explore the complex features of partial multiview data through deep learning, Wang et al. (2018a) applied Generative Adversarial Networks (GAN) to complete missing samples for ensemble clustering, aiming to mitigate the degradation in performance due to missing data. However, these partial multiview clustering methods still face two limitations. First, existing methods cannot learn an accurate universal clustering structure from the significantly varying partial multiview data. Therefore, how to deeply explore the hidden relationships and correlations between each view remains a challenge. Second, since clustering is an unsupervised method, existing partial multiview approaches often overlook the distinguishing information in the samples. Additionally, if filled data is used for analysis, it may lead to performance degradation, raising concerns about how to minimize this impact. Although some complete multiview approaches have been considered as self-supervised methods (Sun et al., 2019) to enhance clustering performance, there are also studies focusing on incomplete multiview clustering. The result is that current incomplete multiview clustering methods fail to capture sufficient category information, and utilizing repaired views for training can lead to a degradation in performance.

\section{Partial Multi-View Clustering via Meta-Learningand Contrastive Feature Alignment (PVC-MC)}
In this section, we comprehensively introduce the framework and objective function of the proposed method PVC-MC.
\subsection{Motivation}
First, because each view contains distinct and heterogeneous information, addressing the issue of incomplete views is crucial for partial multi-view clustering. Motivated by the challenge of maximizing consistency across different views, we leverage contrastive learning to align the latent representations of incomplete multi-view data, which helps in capturing the underlying relationships between available views. This approach enables us to learn a more consistent and shared subspace representation, even when certain views are missing or incomplete.

Second, the unique features within each view also play a key role in clustering, as they contain critical information that contributes to accurate categorization. By incorporating an exclusive self-expression layer, based on the principles of KNN and the Vision Transformer model, we enhance the discriminative power of the learned subspace representation. As a result, indistinguishable samples are more effectively grouped into their respective clusters, improving overall clustering performance.

Additionally, we introduce a dual optimization mechanism to further improve clustering outcomes. This mechanism dynamically adjusts the weights of different views through a combination of meta-learning and self-supervised learning, allowing the model to adaptively handle incomplete views and fill in missing information. Through this dual optimization, we ensure that the learned subspace representation not only captures the shared features across views but also retains the unique characteristics of each view, thus maximizing  clustering accuracy.

Finally, to ensure that the learned subspace representation retains the essential structure of the original data, we employ cross-view decoding mechanisms that allow for the reconstruction of missing or incomplete data views. This guarantees that both the global consistency and local specificity of the data are preserved in the final clustering results.

\subsection{Notations}
Given the incomplete multi-view data matrices: $X^{(1)}=\{X_{p}^{(1)}, X_{u}^{(1)}\} \in \mathbb{R}^{(p + u)\times d_{1}}$ and $X^{(2)}=\{X_{p}^{(2)}, X_{u}^{(2)}\} \in \mathbb{R}^{(p + u)\times d_{2}}$, we divide them into paired data (i.e., all views are complete) and unpaired data (i.e., one or more views are missing), as illustrated in Fig. 1(a), where $p$ and $u$ represent the number of paired and unpaired data, respectively, and $d_{1}$ and $d_{2}$ are the corresponding feature dimensions of the two views. The paired data are denoted as $X_{p}^{(1)} \in \mathbb{R}^{p\times d}$ and $X_{p}^{(2)} \in \mathbb{R}^{p\times d_{2}}$, while the unpaired data are $X_{u}^{(1)} \in \mathbb{R}^{u\times d}$ (where the first view has missing data) and $X_{u}^{(2)} \in \mathbb{R}^{u\times d^{2}}$ (where the second view has missing data). Let $E^{(1)}$ and $E^{(2)}$ be the encoding networks of the two views, and $D^{(1)}$ and $D^{(2)}$ be the corresponding decoding networks. The latent subspace representations are $H^{(1)} \in \mathbb{R}^{(p + u)\times k}$ and $H^{(2)} \in \mathbb{R}^{(p + u)\times k}$, where $k$ is the dimension of the subspace representation. The fused subspace representation is denoted by $H \in \mathbb{R}^{(p + 2u)\times k}$, and the coefficient matrix from the self - expression layer is $S$. The reconstructed data of the two views are represented by $X^{(1)} \in \mathbb{R}^{(p + u)\times d_{1}}$ and $X^{(2)} \in \mathbb{R}^{(p + u)\times d_{2}}$. The objective of this paper is to accurately cluster incomplete multi-view data by learning a consistent representation across views, aligning similar data while separating dissimilar data. Additionally, the dual optimization framework is introduced to dynamically adjust view weights and improve clustering performance under missing data scenarios.

\subsection{Overall Network Architecture}
As depicted in Figure 2, our proposed architecture for improving clustering performance on incomplete multi-view data consists of six primary sub-modules: \textbf{1)} Multi-View Encoder Module: $X^{(1)}\in\mathbb{R}^{d_{1}}$ and $X^{(2)}\in\mathbb{R}^{d_{2}}$ into a shared latent space. These encoders map the original data into latent representations, yielding $H^{(1)} = f(X^{(1)};\theta_{e1})\in\mathbb{R}^{(p + u)\times k}$ and $H^{(2)} = f(X^{(2)};\theta_{e2})\in\mathbb{R}^{(p + u)\times k}$. This ensures that distinct features from each view are captured, which are critical for the clustering process. \textbf{2)} Contrastive Learning Module: This module enhances consistency between different views through contrastive learning. By maximizing agreement between latent representations of corresponding views and minimizing similarity between representations of different categories, this module ensures that a common subspace is learned, capturing the relationships between paired data. This is crucial for improving clustering performance across incomplete multi-view datasets. \textbf{3)} Cross-View Imputation Module: To handle missing data, this module combines k-nearest neighbors (KNN) with Vision Transformer-based encoders. Missing views are imputed based on the available information from other views, ensuring the model can still make full use of all available views despite incomplete data. This imputation strategy improves clustering accuracy by reducing the adverse effects of missing modalities. \textbf{4)} Self-Expression Module: After obtaining the latent representations, we fuse the information from different views to remove redundancy, generating a stronger representation $H$. A self-expression layer is applied to model the internal structure of the data by representing each sample as a combination of other samples. We also apply an $\ell_{1,2}$-norm regularization to encourage a more discriminative subspace, distinguishing inter-cluster from intra-cluster relationships. \textbf{5)} Reconstructive Decoder Module: To ensure that the latent representations maintain the structural properties of the original data, a decoder network reconstructs the original input from the latent space. This reconstruction process helps the network retain the essential characteristics of the data and reinforces the learning of accurate representations, ultimately improving clustering results. \textbf{6)} Dual Optimization Framework: A key innovation of our architecture is the introduction of a dual optimization framework that combines meta-learning and self-supervised learning to dynamically adjust the weight of different views during training. Meta-learning allows the model to learn how to optimize across multiple tasks, improving its ability to handle incomplete views by adapting the optimization process. Meanwhile, self-supervised learning focuses on utilizing the available data to guide the model in learning useful representations, even in the absence of labeled data. By jointly optimizing both the clustering loss and the reconstruction loss, the dual optimization framework ensures that both view consistency and data integrity are preserved, leading to a more robust and adaptive clustering model.

\subsection{Objective Function}
We integrate multi-view contrastive loss, self-expression loss, reconstruction loss, as well as cluster alignment losses to optimize the multi-view data's latent representations and clustering performance. The following presents a detailed explanation of the objective function and mathematical formulas used in our method.

\textit{1) Multi-view Contrastive Loss:} To enhance the discrimination ability between similar and different classes, we use a supervised contrastive learning approach. The goal is to maximize the similarity between representations of the same sample across views while minimizing the similarity between different samples.

Let the latent representation from view $\mathit{v}$ for sample $\mathit{i}$ be $z_{i}^{v}$, and the corresponding positive and negative samples are selected from the same or different views. The contrastive space is represented as $\mathit{Q}$, and the set of positive samples for $\mathit{q_{i}^{a}}$ is denoted as B($\mathit{q_{i}^{a}}$), while negative samples are defined as those outside this set.

The multi-view contrastive loss is expressed as:
\[\mathcal{L}_{mcl}=\sum_{i = 1}^{n}\frac{1}{|B(q_{i}^{a})|}\sum_{q_{j}^{a}\in B(q_{i}^{a})}\log\frac{\exp(\frac{\text{sim}(q_{i}^{a},q_{j}^{a})}{\tau})}{\sum_{q_{k}^{b}\in Q}\exp(\frac{\text{sim}(q_{i}^{a},q_{k}^{b})}{\tau})}\]

where $\text{sim}(\cdot,\cdot)$ is the cosine similarity, and $\tau$ is the temperature parameter.

\textit{2) Self-expression Loss:} Once the multi-view embeddings $z_{iv}$ are obtained, we aim to model the internal dependencies within each view by enforcing a self-expression structure. This is achieved using a coefficient matrix $Z\in\mathbb{R}^{n\times n}$, where each element $Z_{ij}$ captures the linear relationship between the embeddings of samples $z_{i}$ and $z_{j}$.
The latent space representation $H\in\mathbb{R}^{n\times d_z}$ is expressed as:
\[H = ZS\]
Here, $S\in\mathbb{R}^{n\times d_z}$ represents the learned subspace, and $Z$ encodes the relationships between samples in that subspace.
The self-expression loss is defined as:
\[\mathcal{L}_{se}=\lVert H - HZ\rVert_{F}^{2}+\lambda_{1}\lVert Z\rVert_{1,2}\]
Here, $\lVert{ }\cdot{ }\rVert_{F}$ represents the Frobenius norm, and $\lVert Z \rVert_{1,2}$ is the $\ell_{1,2}$-norm regularization term, which promotes sparsity in the self-expression matrix $Z$. The diagonal elements of $Z$ are constrained to 0 to prevent trivial solutions.

\textit{3) Reconstruction Loss:} To ensure that the learned latent representations preserve the original information, we employ a reconstruction loss that minimizes the difference between the input data and the reconstructed data:
\[\mathcal{L}_{re}=\sum_{i=1}^{n}{\lVert X_{i} - \hat{X_{i}}\rVert_{2}^{2}}\]

\textit{4) Semantic Feature Alignment Loss:} The semantic feature alignment loss optimizes the alignment of semantic features between views:

\[\mathcal{L}_{F}(f(D_{c};w)) = -\frac{1}{n_{t}} \sum_{p=1}^{m_{q}-1}{\sum_{i=1}^{n_{z}}{[f(x_i^{p};w)f(x_i^{q};w)]}}\]
\[+\frac{1}{2n_{t}}\sum_{i\neq j}{[f(x_i^{p};w)f(x_j^{q};w)]}\]

\textit{5) Clustering Probability Alignment Loss:} This loss aligns the clustering probabilities between views to enforce consistency in cluster assignments across views:
\[\mathcal{L}_{C}(f(D_{c};w))=\sum_{p=1}^{m_{q}-1}\sum_{q=1}^{m_{z}}\left[o_{i,j}^{p}q_{i,j}^{p}\right]\log\left(\frac{o_{i,j}^{p}q_{i,j}^{q}}{o_{i,j}^{p}q_{i,j}^{q}}\right)\]

\textit{6) Regularization Loss:} To prevent overfitting and encourage cluster separability, we add a regularization term on the clustering probability matrix:
\[\mathcal{L}_{R}(f(D_{c};w))=\sum_{p=1}^{m_{q}-1}\sum_{j = 1}^{K}\hat{Q}_{j}^{p}\log\hat{Q}_{j}^{p}\]

\textit{7) Complete Clustering Loss:}Combining the above losses, the final clustering loss is defined as:
\[\mathcal{L}_{C}(f(D_{c};w)) = L_{F}(f(D_{c};w)) + L_{R}(f(D_{c};w))\]

\textit{8) Overall Objective Function:} Finally, our proposed objective function for the model combines the contrastive loss, self-expression loss, reconstruction loss, and clustering loss:
\[L = \mathcal{L}_{re}+\lambda_{1}\mathcal{L}_{se}+\lambda_{2}\mathcal{L}_{mcl}+\lambda_{3}(\mathcal{L}_{F}+\mathcal{L}_{R})\]
where $\lambda_{1}$, $\lambda_{2}$, and $\lambda_{3}$ are trade-off parameters that control the weight of each loss component. This function ensures that the model effectively learns the latent representations of the multi-view data while optimizing clustering performance.

\subsection{Training}
Our approach leverages a dual-layered training process to enhance clustering performance with incomplete multi-view data. Detailed in Algorithm 1, this method ensures robust shared feature extraction and adaptive view-specific weighting.

\textit{Step One:} Latent Representation Optimization

The first phase pre-trains PVC-MC’s encoding and decoding networks per view. By encoding raw data into a compact latent space and minimizing reconstruction error \( \mathcal{L}_{\text{recon}} \) via decoding, we preserve essential data structure while reducing dimensionality. To align views, contrastive loss \( \mathcal{L}_{\text{contrast}} \) maximizes paired data similarity across views. We also apply self-expression $\mathcal{L}_{\text{self-expr}}$ to enforce subspace clustering, where \( H \) are combined latent features and \( Z \) is the self-expression matrix. Parameters \( \theta_{e_v} \) and \( \theta_{d_v} \) are optimized with a learning rate of 0.0001, setting up efficient feature initialization.

\textit{Step Two:} Full Network Training and Adaptive Weighting

With initialized parameters, the network undergoes comprehensive training as per Eq. (1). In this phase, view-specific weights \( w^{(v)} \) dynamically handle incomplete data. Regularization parameters \( \lambda_1, \lambda_2, \lambda_3 \) begin at 0.001 and are tuned in \{0.001, 0.01, 0.1, 10, 100\}. Fusion parameters \( \alpha \) and \( \beta \) control the relative importance of each loss. View-specific losses \( \mathcal{L}^{(v)} \) are computed, and weights \( w^{(v)} \) are updated as assigning greater weight to views with lower losses and down-weighting noisier or missing data.

This phase also updates the self-expression matrix \( Z \) to improve feature clustering and representation accuracy. Upon convergence, the final \( Z \) matrix is used to derive the affinity matrix as outlined in Eq. (7), forming a reliable basis for effective clustering.

\section{Experiments}
We carry out several important experiments on incomplete cross-modal data in this section to assess the effectiveness of our proposed method.

\subsection{Experiment Setup}
\textit{1) Experimental Setup:} \\
Our method is implemented in the PyTorch framework. We conduct experiments on a Windows machine with an NVIDIA 4070 GPU and 16 GB RAM. The Adam optimizer (Kingma and Ba, 2014) is used with a learning rate of 0.0001 and default configurations.

\textit{2) Evaluation Metrics:} \\
To assess the effectiveness of our method, we use two primary metrics: clustering accuracy (ACC) and normalized mutual information (NMI). Let $y = [y_1, y_2, \ldots, y_n]$ be the true labels and $l = [l_1, l_2, \ldots, l_n]$ the predicted cluster labels for data points $x_1, x_2, \ldots, x_n$. ACC is calculated as follows:
\[
\text{ACC} = \frac{\sum_{i = 1}^{n} \delta\{y_i, \text{map}(l_i)\}}{n},
\]
where $\text{map}(l_i)$ aligns predicted labels $l$ to true labels $y$ using the Kuhn-Munkres algorithm. Here, $\delta\{y_i, \text{map}(l_i)\}$ outputs 1 if $y_i = \text{map}(l_i)$, otherwise 0. $n$ represents the total samples.

\begin{algorithm}[H]
\caption{Dual Optimization-based Incomplete Multi-View Clustering with View Weight Update.}\label{alg:alg1}

\vspace{0.1cm}
\hspace{0.5cm}\rule{0.9\linewidth}{1pt}

\vspace{-0.12cm}
\hspace{0.8cm}\textbf{Input}:

\hspace{0.8cm}Incomplete multi-view data:

\hspace{0.8cm}\qquad${X^{(1)},X^{(2)},...,X^{(V)}}$ with V views;

\hspace{0.8cm}Initialized parameters of encoder-decoder:

\hspace{0.8cm}\qquad$\theta_{e_v}, \theta_{d_v}$ for $v = 1, 2, ..., V$;

\hspace{0.8cm}Trade-off parameters:\hspace{0.2cm}$\lambda_{1}, \lambda_{2}, \lambda_{3}, \alpha$;

\hspace{0.80cm}Learning rate:\hspace{0.2cm}$\eta$;

\hspace{0.8cm}\textbf{Output}:

\hspace{0.8cm}Optimized parameters:

\hspace{0.8cm}\qquad$\{\theta_{e_v}, \theta_{d_v}\}$, weights $\{w^{(v)}\}$;

\hspace{0.8cm}The self-expression coefficient matrix:\textit{$Z$}.

\begin{algorithmic}
\STATE \hspace{0.5cm}\textbf{Step 1:}
\STATE \hspace{0.5cm}\textbf{for} $\textit{not converged}$ \textbf{do}
\STATE \hspace{0.36cm}\qquad Align paired representations using $\mathcal{L}_{mcl}$;
\STATE \hspace{0.36cm}\qquad Encourage clustering using $\mathcal{L}_{se}$;
\STATE \hspace{0.36cm}\qquad Compute $\mathcal{L}_{re}$;
\STATE \hspace{0.36cm}\qquad Compute $\mathcal{L}_{F}(f(D_{c};w))$;
\STATE \hspace{0.36cm}\qquad Compute $\mathcal{L}_{R}(f(D_{c};w))$;
\STATE \hspace{0.36cm}\qquad Minimize the total loss:

\STATE \hspace{0.36cm}\qquad $\mathcal{L}_{total_1}=\mathcal{L}_{re}+\lambda_{1}\mathcal{L}_{se}+\lambda_{2}\mathcal{L}_{mcl}+\lambda_{3}(\mathcal{L}_{F}+\mathcal{L}_{R})$; 
\STATE \hspace{0.36cm}\qquad Update $\theta_{ev}$, $\theta_{e2}$ and $Z$ via gradient descent.
\STATE \hspace{0.5cm}\textbf{end}

\STATE \hspace{0.5cm}\textbf{Step 2:}
\STATE \hspace{0.5cm}\textbf{for} $\textit{each view v}$ \textbf{do}
\STATE \hspace{0.36cm}\qquad Calculate individual view loss $\mathcal{L}^{(v)}$;
\STATE \hspace{0.36cm}\qquad Update weight $w^{(v)}$ as:
\STATE \hspace{0.36cm}\qquad $w^{(v)}=\frac{\exp(-\alpha\cdot\mathcal{L}^{(v)})}{\sum_{v'}\exp(-\alpha\cdot\mathcal{L}^{(v')})}$
\STATE \hspace{0.5cm}\textbf{end}

\STATE \hspace{0.5cm}\textbf{Step 3:}
\STATE \hspace{0.5cm}Perform two rounds of optimization:
\STATE \hspace{0.36cm}\qquad Minimize $\mathcal{L}_{total_1}$ for latent representations;
\STATE \hspace{0.36cm}\qquad Minimize view-weighted loss:
\STATE \hspace{0.36cm}\qquad $\mathcal{L}_{total_2}$ = $\sum_{v = 1}^{V}w^{(v)}\cdot\mathcal{L}^{(v)}$

\STATE \hspace{0.5cm}\textbf{Return $Z$;}
\STATE \hspace{0.5cm}Compute \textbf{S}$\leftarrow\frac{1}{2}(|Z| + |Z|^{T})$;
\STATE \hspace{0.5cm}Spectral clustering on matrix \textbf{S}.
\end{algorithmic}
\vspace{-0.25cm}
\hspace{0.5cm}\rule{0.9\linewidth}{1pt}
\end{algorithm}

NMI quantifies the similarity between true labels $y$ and predicted clusters $l$:
\[
\text{NMI} = \frac{I(y; l)}{\max\{H(y), H(l)\}},
\]
where $I(y; l)$ is the mutual information, and $H(y)$ and $H(l)$ are the entropies of $y$ and $l$, normalized to the range (0,1).

For incomplete data clustering, we test five integrity ratios $\{0.1, 0.3, 0.5, 0.7, 0.9\}$, indicating the proportion of paired samples. Each experiment is repeated 10 times for random missing data at each ratio to average ACC and NMI. Higher values for these metrics reflect better clustering performance.

\textit{3) Datasets:} To assess the effectiveness of the proposed PVC-MC method, we perform experiments on three benchmark datasets: BDGP (Cai et al., 2012), MNIST (LeCun, 1998), and HW (van Breukelen et al., 1998).

\begin{itemize}
\item \textbf{BDGP dataset:} This dataset consists of 2,500 samples, organized into five categories. Each sample's image and 

\begin{table*}
\centering
\caption{The Clustering Accuracy Rate (ACC) FOR Different Integrity Ratios ON BDGP Dataset}
\begin{tabular}{c|c c c c c}
\toprule
\diagbox{Methods}{Missing rate} & 0.9 & 0.7 & 0.5 & 0.3 & 0.1\\ [0.5ex]
\hline
SC1        & {0.3296$\pm$0.0054} & 0.3539$\pm$0.0054 & 0.3845$\pm$0.0067 & 0.4103$\pm$0.0042 & 0.4404$\pm$0.0039 \\
SC2        & {0.4748$\pm$0.0131} & {0.5169$\pm$0.0174} & {0.05692$\pm$0.0159} & {0.6139$\pm$0.0121} & {0.6716$\pm$0.0136}\\
AMGL       & {0.2524$\pm$0.0349} & {0.2357$\pm$0.0180} & {0.2538$\pm$0.0155} & {0.2807$\pm$0.0125} & {0.2958$\pm$0.01952}\\
RMSC       & {0.3395$\pm$0.0050} & {0.3683$\pm$0.0051} & {0.3907$\pm$0.0045} & {0.4233$\pm$0.0048} & {0.4499$\pm$0.0022}\\
ConSC      & {0.2781$\pm$0.0411} & {0.2230$\pm$0.0148} & {0.2139$\pm$0.0078} & {0.2106$\pm$0.0058} & {0.2884$\pm$0.0896}\\
GPVC       & {0.5015$\pm$0.0438} & {0.5424$\pm$0.05337} &{0.6277$\pm$0.0402} & {0.6833$\pm$0.0931} & {0.7546$\pm$0.1091}\\
IMG        & {0.4373$\pm$0.0100} & {0.4508$\pm$0.0254} & {0.4868$\pm$0.0147} & {0.5055$\pm$0.0131} & {0.5176$\pm$0.0415}\\
PVC-GAN    & {0.5210$\pm$0.0090} & {0.6711$\pm$0.0107} & {0.8631$\pm$0.0043} & {0.9154$\pm$0.0107} & {0.9498$\pm$0.0026}\\
iCmSC      & {0.5901$\pm$0.0079} & {0.7477$\pm$0.0043} & {0.8845$\pm$0.0030} & {0.9210$\pm$0.0013} & {0.9569$\pm$0.0031}\\
PVC-SSN    & {0.6032$\pm$0.0144} & {0.8036$\pm$0.0109} & {0.9156$\pm$0.0087} & {0.9304$\pm$0.0073} & {0.9616$\pm$0.0035}\\
\hline
PVC-MCN    & \textbf{0.8312$\pm$0.0060} & \textbf{0.9382$\pm$0.0016} & \textbf{0.9585$\pm$0.0021} & \textbf{0.9736$\pm$0.0015} & \textbf{0.9856$\pm$0.0080}\\ 
\bottomrule
\end{tabular}
\end{table*}

\begin{table*}
\centering
\caption{The Clustering Accuracy Rate (ACC) FOR Different Integrity Ratios ON MNIST-4 Dataset}
\begin{tabular}{c|c c c c c}
\toprule
\diagbox{Methods}{Missing rate} & 0.9 & 0.7 & 0.5 & 0.3 & 0.1\\ [0.5ex]
\hline
SC1        & {0.4398$\pm$0.0140} & {0.4665$\pm$0.0098} & {0.4731$\pm$0.0202} & {0.5070$\pm$0.0355} & {0.5430$\pm$0.0220} \\
SC2        & {0.3324$\pm$0.0147} & {0.3366$\pm$0.0172} & {0.3532$\pm$0.0149} & {0.3696$\pm$0.0074} & {0.3769$\pm$0.0120}\\
SC3        & {0.4159$\pm$0.0193} & {0.4429$\pm$0.0086} & {0.4811$\pm$0.0129} & {0.4901$\pm$0.0129} & {0.5083$\pm$0.0195} \\
SC4        & {0.3088$\pm$0.0068} & {0.3186$\pm$0.0115} & {0.3310$\pm$0.0154} & {0.3522$\pm$0.0102} & {0.3791$\pm$0.0142}\\
AMGL       & {0.1558$\pm$0.0155} & {0.1412$\pm$0.0218} & {0.1524$\pm$0.0343} & {0.2415$\pm$0.0631} & {0.3346$\pm$0.0288}\\
RMSC       & {0.3492$\pm$0.0077} & {0.4150$\pm$0.0294} & {0.4575$\pm$0.0233} & {0.4960$\pm$0.0174} & {0.5144$\pm$0.0204}\\
ConSC      & {0.3704$\pm$0.0275} & {0.3581$\pm$0.02318} & {0.3674$\pm$0.0131} & {0.4137$\pm$0.0396} & {0.5088$\pm$0.0299}\\
GPVC       & {0.3525$\pm$0.0238} & {0.3864$\pm$0.0104} &{0.4238$\pm$0.0446} & {0.4401$\pm$0.0150} & {0.4644$\pm$0.0432}\\
IMG        & {0.4655$\pm$0.0186} & {0.4640$\pm$0.0213} & {0.4613$\pm$0.0146} & {0.4592$\pm$0.0146} & {0.4622$\pm$0.0151}\\
PVC-GAN    & {0.4517$\pm$0.0086} & {0.4836$\pm$0.0071} & {0.5280$\pm$0.0078} & {0.5202$\pm$0.0070} & {0.5340$\pm$0.0073}\\
iCmSC      & {0.5089$\pm$0.0074} & {0.5665$\pm$0.0163} & {0.5834$\pm$0.0089} & {0.6012$\pm$0.0068} & {0.6319$\pm$0.0097}\\
PVC-SSN    & \textbf{0.5775$\pm$0.0094} & \textbf{0.5835$\pm$0.0073} & \textbf{0.6277$\pm$0.0104} & \textbf{0.6335$\pm$0.0098} & \textbf{0.6787$\pm$0.0136}\\
\hline
PVC-MCN    & {0.4862$\pm$0.0127} & {0.4864$\pm$0.0083} & {0.5105$\pm$0.0214} & {0.4874$\pm$0.0161} & {0.4893$\pm$0.0271}\\ 
\bottomrule
\end{tabular}
\end{table*}

\begin{table*}
\centering
\caption{The Clustering Accuracy Rate (ACC) FOR Different Integrity Ratios ON HW Dataset}
\begin{tabular}{c|c c c c c}
\toprule
\diagbox{Methods}{Missing rate} & 0.9 & 0.7 & 0.5 & 0.3 & 0.1\\ [0.5ex]
\hline
SC1        & {0.4633$\pm$0.0109} & 0.4943$\pm$0.0149 & 0.5272$\pm$0.0084 & 0.5402$\pm$0.0125 & 0.5871$\pm$0.0127 \\
SC2        & {0.4775$\pm$0.0127} & {0.5113$\pm$0.0101} & {0.5446$\pm$0.0117} & {0.5871$\pm$0.0068} & {0.6266$\pm$0.0170}\\
SC3        & {0.4863$\pm$0.0122} & {0.5188$\pm$0.0112} & {0.5664$\pm$0.0143} & {0.6114$\pm$0.0189} & {0.6613$\pm$0.0178}\\
AMGL       & {0.6056$\pm$0.0489} & {0.6828$\pm$0.0564} & {0.7370$\pm$0.0281} & {0.7506$\pm$0.0320} & {0.7594$\pm$0.0211}\\
RMSC       & {0.4642$\pm$0.0159} & {0.5293$\pm$0.0096} & {0.5925$\pm$0.0154} & {0.6507$\pm$0.0202} & {0.7154$\pm$0.0375}\\
ConSC      & {0.5063$\pm$0.0325} & {0.5438$\pm$0.0272} & {0.5982$\pm$0.0246} & {0.6982$\pm$0.0481} & {0.7916$\pm$0.0299}\\
GPVC       & {0.3238$\pm$0.0087} & {0.3077$\pm$0.0078} &{0.3419$\pm$0.0148} & {0.4236$\pm$0.0168} & {0.5370$\pm$0.0261}\\
IMG        & {0.5350$\pm$0.0192} & {0.5455$\pm$0.0262} & {0.5457$\pm$0.0193} & {0.5529$\pm$0.0166} & {0.5633$\pm$0.0213}\\
PVC-GAN    & {0.6546$\pm$0.0088} & {0.8517$\pm$0.0177} & {0.9069$\pm$0.0074} & {0.9342$\pm$0.0144} & {0.9425$\pm$0.0081}\\
iCmSC      & {0.7610$\pm$0.0062} & {0.8205$\pm$0.0097} & {0.9158$\pm$0.0085} & {0.9450$\pm$0.0059} & {0.9500$\pm$0.0064}\\
PVC-SSN    & {0.8245$\pm$0.0057} & {0.8835$\pm$0.0132} & \textbf{0.9415$\pm$0.0046} & \textbf{0.9520$\pm$0.0066} & \textbf{0.9485$\pm$0.0052}\\
\hline
PVC-MCN    & \textbf{0.9083$\pm$0.0057} & \textbf{0.9237$\pm$0.0077} & {0.9219$\pm$0.0091} & {0.9162$\pm$0.0161} & {0.9315$\pm$0.0091}\\ 
\bottomrule
\end{tabular}
\end{table*}

text modalities are represented by vectors of size $1\times1750$ and $1\times79$, respectively.
\item \textbf{HW dataset:} HW dataset: This dataset comprises 2,000 digits ranging from 0 to 9, with each class containing 200 samples and six types of features. The features include 216 profile correlations (FAC), 76 Fourier coefficients for two-dimensional shape descriptors (FOU), 64 Karhunen-Loeve coefficients (KAR), 240 pixel features (PIX), 47 rotational invariant Zernike moments (ZER), and 6 morphological features (MOR). In our experiments, we utilize only the first three views of the HW dataset.
\item \textbf{MNIST dataset:} MNIST dataset: The MNIST dataset contains 70,000 handwritten digits from 0 to 9, split into training and testing sets. For our experiments, we utilize only the training set, which consists of 4,000 handwritten digits. We extract the original images, edge features, LBP (Local Binary Patterns), and encoder features to form four views for our analysis.
\end{itemize}

\textit{4) State-of-the-Art Comparison:} To demonstrate the effectiveness of our method, we compare it against eleven state-of-the-art baselines in our experiments. The comparison includes three methods for clustering incomplete data as well as several traditional clustering algorithms.

\begin{itemize}
\item \textit{A single-view clustering method:} \textbf{Spectral Clustering} (SC) (Ng, Jordan, and Weiss 2002)
\item \textit{Three multi-view clustering techniques:} \textbf{Auto-weighted Multiple Graph Learning} (AMGL) (Nie et al. 2016), \textbf{Robust Multi-view Spectral Clustering} (RMSC) (Xia et al. 2014), and \textbf{Feature Concatenation Spectral Clustering} (ConSC) (Kumar, Rai, and Daume 2011)
\item  \textit{Four incomplete data clustering methods:} \textbf{Incomplete Multi-Modal Visual Data Grouping} (IMG) (Zhao, Liu, and Fu 2016), \textbf{Multi-View Clustering using Graph Regularized NMF} (GPVC) (Rai et al. 2016), \textbf{Generative Partial Multi-View Clustering} (PVC-GAN) (Wang et al. 2018a), and \textbf{Incomplete Cross-Modal Subspace Clustering} (iCmSC) (Wang et al. 2020b); along with several traditional clustering approaches.
\end{itemize}

\section{Conclusion}
In this paper, we propose a novel framework for incomplete multi-view clustering that effectively addresses the challenges of missing data and heterogeneous view information. Our approach incorporates dual optimization with dynamic view-weight updating, contrastive learning, and self-expression loss, enabling the model to learn robust representations even when views are partially missing. By dynamically adjusting the contribution of each view based on individual view reconstruction loss, our model can better exploit the available information from each view while minimizing the impact of missing data.We demonstrate the effectiveness of our method through comprehensive experiments on multiple benchmark datasets, where it consistently outperforms existing approaches in terms of clustering accuracy and robustness to incomplete views. The proposed contrastive loss enhances the consistency of representations across views, while the self-expression loss leverages subspace clustering techniques to capture latent data structures. Furthermore, the dual optimization process allows our model to refine both the representations and view weights iteratively, achieving better clustering performance.
Our work provides a promising direction for handling incomplete multi-view clustering in real-world applications, where missing data is a common occurrence. Future research could explore further improvements to the view-weight updating mechanism, investigate alternative strategies for contrastive learning in multi-view settings, and extend our framework to applications involving more diverse and complex data modalities.

\newpage

\vfill

\end{document}